\documentclass[10pt,twocolumn,letterpaper]{article}

\usepackage{cvpr}
\usepackage{times}
\usepackage{epsfig}
\usepackage{graphicx}
\usepackage{amsmath}
\usepackage{amssymb}
\usepackage{makecell}
\usepackage{multirow}

\usepackage{mathtools}
\usepackage{amsmath}
\usepackage{amssymb}
\usepackage{amsfonts}
\usepackage{amsopn}
\usepackage{graphicx} % Necessary to use \scalebox
\usepackage{textcomp}
\usepackage{xfrac}
\usepackage{bbm}
\usepackage{cite}
\usepackage{overpic}
\usepackage{subfig}
\usepackage{wrapfig}
\usepackage[ruled,vlined,linesnumbered]{algorithm2e}
\usepackage{multirow}

\usepackage{color}
\definecolor{green}{rgb}{0, 0.5, 0}
\definecolor{orange}{rgb}{0.8, 0.6, 0.2}
\definecolor{red}{rgb}{1.0, 0.0, 0.0}
\definecolor{teal}{rgb}{0.0, 0.4, 0.4}
\definecolor{purple}{rgb}{0.65,0,0.65}
\definecolor{saffron}{rgb}{0.95,0.75,0.2}
\definecolor{turquoise}{rgb}{0.0,0.5,0.5}
\definecolor{brown}{rgb}{0.5, 0.16, 0.16}

\usepackage{overpic}
\usepackage{currfile} % \currfiledir

\newcommand{\myfigurename}{\put(-3,0){\vertical{\todo{\currfiledir}}}}
\renewcommand{\myfigurename}{}

\newcommand{\supl}[1]{{\color{black}#1}}

\newcommand{\kx}[1]{{\color{black}#1}}

\definecolor{lightgray}{rgb}{0.6, 0.6, 0.6}

\newcommand{\Fig}[1]{Figure~\ref{fig:#1}}

\usepackage[normalem]{ulem}

\renewcommand{\paragraph}[1]{\textbf{#1.}}

\newcommand{\hidecomment}[1]{}
\usepackage{xspace}
\cvprfinalcopy % *** Uncomment this line for the final submission

\def\cvprPaperID{6820} % *** Enter the CVPR Paper ID here

\ifcvprfinal\fi
\begin{document}

%%%%%%%%% TITLE
%\title{Pose from Shape: Category-Level 6D Object Pose Estimation\\via Normalized Point Cloud Reconstruction}
%\title{Learning View-Factorized Shape Embedding for Category-Level\\6D Object Pose Estimation}
%\title{View-Factorized Shape Space for Category-Level 6D Object Pose Estimation}
\title{Learning Canonical Shape Space for Category-Level\\6D Object Pose and Size Estimation}

\author{
Dengsheng Chen$^{1,\ast}$ \quad\quad Jun Li$^{1,}$\thanks{Joint first authors} \quad\quad Zheng Wang$^{2}$
\quad\quad Kai Xu$^{1,}$\thanks{Corresponding author: kevin.kai.xu@gmail.com}\\
$^1$National University of Defense Technology \quad $^2$Taobao.com\\
}

\maketitle
%\thispagestyle{empty}

% -*- compile-command: "texify --pdf --quiet decop_cvpr18.tex" -*-
% !TEX root = decop_cvpr18.tex
% (shell-command "start decop_cvpr18.pdf")

\begin{abstract}
We present a novel approach to category-level 6D object pose and size estimation. To tackle intra-class shape variations, we learn canonical shape space (CASS), a unified representation for a large variety of instances of a certain object category. In particular, CASS is modeled as the latent space of a deep generative model of canonical 3D shapes with normalized pose. We train a variational auto-encoder (VAE) for generating 3D point clouds in the canonical space from an RGBD image. The VAE is trained in a cross-category fashion, exploiting the publicly available large 3D shape repositories. Since the 3D point cloud is generated in normalized pose (with actual size), the encoder of the VAE learns view-factorized RGBD embedding. It maps an RGBD image in arbitrary view into a pose-independent 3D shape representation. Object pose is then estimated via contrasting it with a pose-dependent feature of the input RGBD extracted with a separate deep neural networks. We integrate the learning of CASS and pose and size estimation into an end-to-end trainable network, achieving the state-of-the-art performance.
\end{abstract}\vspace{-12pt}

% -*- compile-command: "texify --pdf --quiet decop_cvpr18.tex" -*-
% !TEX root = decop_cvpr18.tex
% (shell-command "start decop_cvpr18.pdf")

\section{Introduction}
6D object pose estimation based on a single-view RGB(D) image is an essential building block
for several real-world applications ranging from robotic navigation and manipulation to augmented reality.
Most existing works have so far been addressing \emph{instance-level} 6D pose estimation where each target
object has a corresponding CAD model with exact shape and size~\cite{peng2019pvnet}.
Thereby, the problem is largely reduced to finding sparse or dense correspondence
between the target object and the stock 3D model.
Pose hypotheses can then be generated and verified based on the correspondences.
Although enjoying high pose accuracy, the requirement of exact CAD models by these techniques hinders their practical use in many application scenarios.

Recently, \emph{category-level} 6D object pose estimation starts to gain attention~\cite{sahin2018category,wang2019normalized}.
In this problem, the target object of a shape category is unseen before and no CAD model is available, although some other instances of the same category may have been seen.
Therefore, the major challenge is how to deal with \emph{intra-class variation}~\cite{sahin2019instance}.
In general, household objects could exhibit significant variations in color, texture, shape and size even within the same category.
%Without a corresponding CAD model, correspondence-based approach does not directly apply.
%There is not a unified representation for various instances of an object category to which the target object in observation could be matched.
Without an exactly same CAD model, correspondence-based approach would find difficulty given the considerable intra-class shape variation.

%\begin{figure}[t]
%	\begin{center}
%		\includegraphics[width=\linewidth]{fig/teaser.pdf}
%	\end{center}
%    \vspace{-8pt}
%	\caption{We present a method for category-level 6D object pose and size estimation (left) via learning canonical shape space. The input RGBD image is embedded into the shape space, resulting in a view-factorized RGBD embedding. Object pose is then estimated via contrasting it with a pose-dependent feature of the input RGBD. A side-product of our method is the full-shape reconstruction of the input single-view RGBD, which supports not only size calculation but also precise robotic grasping. In the figure, the reconstructed 3D point clouds are placed into the scene point cloud (unprojected from the input depth map). Two of them are highlighted in the middle.}
%    \vspace{-6pt}
%	\label{fig:teaser}
%\end{figure}
\begin{figure}[t]
\centering
\begin{overpic}
[width=\linewidth]
%[width=\linewidth,grid,tics=10]
{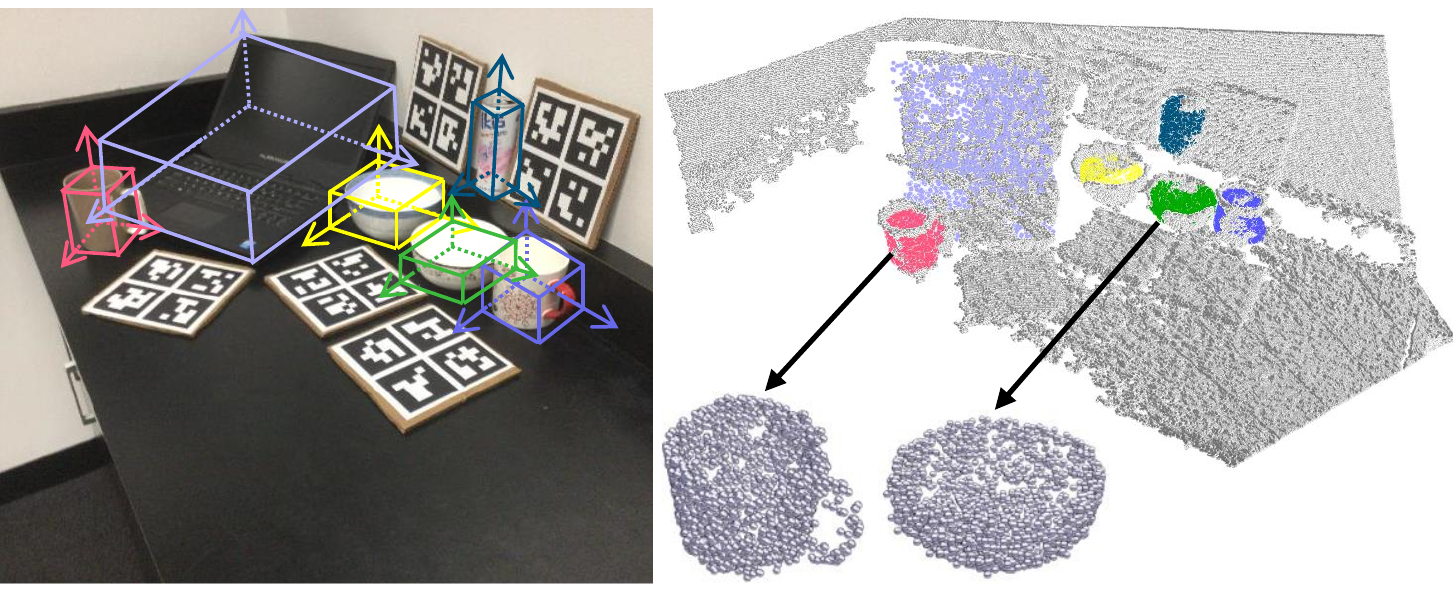}\myfigurename{}
    %\put(17,-3){\small (a)}
    %\put(56,-3){\small (b)}
    %\put(86,-3){\small (c)}
\end{overpic}
\vspace{-8pt}
\caption{
We present a method for category-level 6D object pose and size estimation (left) via learning canonical shape space. The input RGBD image is embedded into the shape space, resulting in a view-factorized RGBD embedding. Object pose is then estimated via contrasting it with a pose-dependent feature of the input RGBD. A side-product of our method is the full-shape reconstruction of the input single-view RGBD, which supports not only size calculation but also precise robotic grasping. In the figure, the reconstructed 3D point clouds are placed into the scene point cloud (unprojected from the input depth map). Two of them are highlighted in the middle.
}
\label{fig:teaser}\vspace{-6pt}
\end{figure}

To resolve this, a \emph{unified representation} for a variety of instances of an object category is needed, to which the target object in observation could be ``matched''.
The recently proposed Normalized Object Coordinate Space (NOCS)~\cite{wang2019normalized} is a nice example of such unified representation. Based on this representation, category-level object poses can be estimated with high accuracy through mapping each pixel of the input image to a point in NOCS.
%A  unified representation, however, can hardly accommodate large amount of shape variants.
Finding \emph{dense} mappings between NOCS and an \emph{unseen} object, however, is an ill-posed problem under significant shape variation. Therefore, a generalizable mapping function accommodating large amount of unknown shape variants can be difficult to learn. %This could result in suboptimal poses as shown in Section~\ref{}.

%How to handle significant variation? We use learned shape space with multi-modal distributions.
In this work, we propose to learn a \emph{canonical shape space (CASS)} as our unified representation.
\kx{CASS is modeled by the latent space of a deep generative model of canonical 3D shapes with normalized pose and actual metric size. In particular, we train a variational auto-encoder (VAE) for generating 3D point clouds in the canonical space from an RGBD image.
The VAE is trained in a cross-category fashion, exploiting the publicly available large 3D shape repositories.
Since the 3D point cloud is generated with normalized pose and metric size,
the encoder of the VAE learns \emph{view-factorized} RGBD embedding.
It maps an RGBD image in arbitrary view into a \emph{pose-independent} 3D shape representation.
Object pose can then be estimated via contrasting it with a \emph{pose-dependent} feature of the input RGBD extracted with a separate deep neural networks (Figure~\ref{fig:teaser}). This circumvents the difficulty in estimating dense correspondence between two representations as in other methods~\cite{peng2019pvnet,wang2019normalized}.}

We integrate the learning of CASS and pose and size estimation into an end-to-end trainable network which involves several key designs.
\emph{First}, through learning the canonical shape space with plentiful shape variants, we obtain a unified representation encompassing adequate shape variations.
\emph{Second}, to overcome the lack of real-world training images with 3D point clouds, we enhance the encoder of the VAE to take both RGBD images and 3D shapes as input. This allows us to train the VAE through exploiting off-the-shelf 3D shape repositories.
\kx{
\emph{Third}, in realizing pose estimation, we opt for feature contrasting over dense correspondence, leading to better generality to unseen instances.
Meanwhile, to match the distributions of pose-dependent and pose-independent features so that pose estimation can be easily trained, we propose a few crucial designs, e.g., network weight sharing and training batch mixing.
\emph{Last}, our VAE model is able to reconstruct a 3D point cloud of the target object with metric size, which reduces the learning difficulty by decoupling the estimation of pose and size.}

%key insight: factorize shape and pose

Through evaluating on public category-level datasets, we show that our method archives the state-of-the-art pose accuracy and comparably high size accuracy.
%Our method also demonstrates the best performance even on the instance-level benchmarks.
%
Our work makes the following contributions:
\begin{itemize}
  \vspace{-6pt}\item We propose a novel correspondence-free approach to category-level object pose and size estimation based on learned canonical shape space.
  \vspace{-6pt}\item We design an end-to-end trainable deep neural network for jointly learning the canonical shape space and estimating object pose and size.
  \vspace{-6pt}\item We devise several key designs to ease the network training such as distribution matching between pose-dependent and pose-independent features.
\end{itemize}

\section{Related Work}

\paragraph{Instance-level approaches}
Many works on instance-level 6D pose estimation adopt template-based methods~\cite{hodavn2015detection,konishi2018real}.
In these methods, a set of RGB(D) templates rendered from CAD models in various poses
are matched against the input image in a sliding-window fashion,
based on hand-designed or learned feature descriptors.
The final pose is retrieved from the best matched template or estimated by 3D model registration.
Another group of works pursue to match the target object to the corresponding 3D model.
Depending on the input modality, the core task is to find 2D-to-3D~\cite{izadinia2017im2cad,georgakis2018matching}, 2.5D(depth)-to-3D~\cite{choi20123d,choi2012voting} or 3D-to-3D correspondence~\cite{avetisyan2019scan2cad}.
Several other works opt to learn a 6D object pose regressor directly from the feature descriptions~\cite{tejani2014latent,sahin2016iterative}.
Brachmann et al.~\cite{brachmann2014learning} learn to regress an object coordinate representation which can then be used in pose estimation.

Learning effective feature representation using convolutional neural networks (CNNs) for robust matching has become a main focus of the recent literature~\cite{wohlhart2015learning,kehl2016deep,balntas2017pose,sock2018multi,zhang2019holistic}.
Another line of works utilize CNNs to detect feature points or corner points~\cite{rad2017bb8,tekin2018real}.
PVNet~\cite{peng2019pvnet} is a unique approach of feature point detection using CNNs: A vector field is estimated for the input RGB image based on which the feature points are voted.
Some other works choose to learn an end-to-end deep model that can directly regress 6D object pose from the raw RGB(D) input~\cite{xiang2017posecnn,do2018deep,li2018deepim}.
SSD-6D~\cite{kehl2017ssd} combines single-shot object detection in RGB images with pose hypothesis regression and verification. Similar approach has also been used for multi-view active pose estimation~\cite{sock2017multi}.
Wang et al.~\cite{wang2019densefusion} propose DenseFusion to learn pixel-wise feature extraction and pose estimation. The network can also predict a confidence for each pose hypothesis for final pose selection.

\kx{
Xiao et al.~\cite{xiao2019pose} train a CNN that takes as input both an image and a CAD model, and outputs object pose with respect to the 3D model. This model can generalize to the target objects which are unseen during training. However, they still require the target CAD model during inference. Thus, we classify it as an instance-level work.
}

%Generally, the performance of end-to-end approaches still cannot match that of feature matching based ones.

\begin{figure*}[t]
\centering
\begin{overpic}
[width=0.98\linewidth]
%[width=\linewidth,grid,tics=10]
{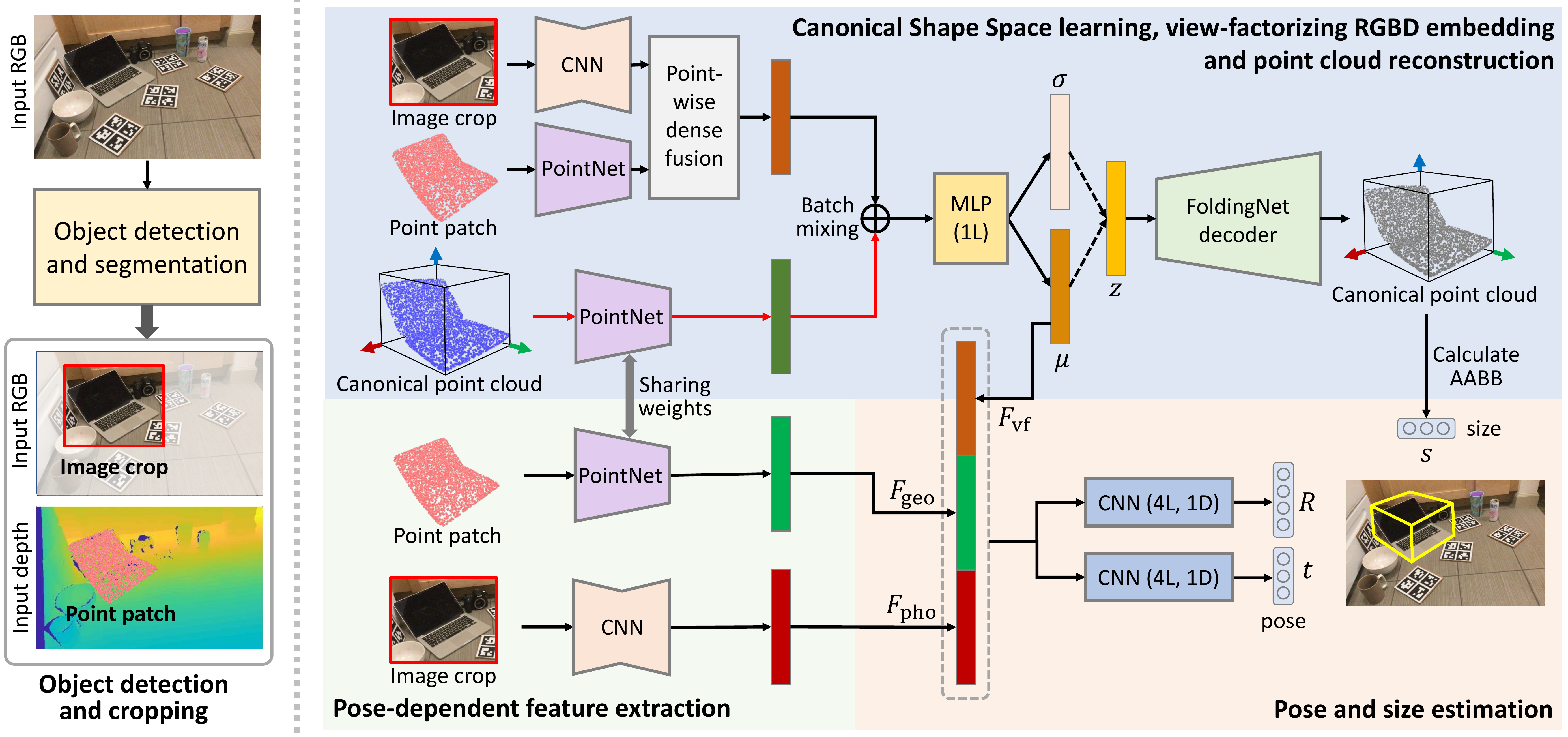}\myfigurename{}
    %\put(17,-3){\small (a)}
    %\put(56,-3){\small (b)}
    %\put(86,-3){\small (c)}
\end{overpic}
\caption{
An overview of our network architecture. A pre-processing (left) is devised to produce image crop and point patch of the object of interest which are fed into the main network (right).
The main network is composed of three modules:
1) CASS learning, pose-factorizing embedding and point cloud reconstruction (background shaded in light blue),
2) pose-dependent feature extraction (light green), and
3) pose estimation (light red).
The network branch indicated with red arrows is used only in training.
}
\label{fig:overview}\vspace{-10pt}
\end{figure*}

\paragraph{Category-level approaches}
There has been a large body of works on category-level object detection and 2D/3D/4D pose estimation.
However, methods designed for estimating \emph{6D poses} is still scarce~\cite{sahin2019instance}.
Sahin et al.~\cite{sahin2018category} introduce a part-based random forest approach for this task.
In their method, parts extracted from CAD model instances of some category are represented with skeletons, which are fed into a random forest for hypothesizing 6D poses. Due to the reliance on purely geometric features, this method mainly deals with depth input.
Wang et al.~\cite{wang2019normalized} introduce Normalized Object Coordinate Space (NOCS) as a shared canonical representation of object instances within a category. They train a region-based neural network to directly infer the pixel-wise correspondence between an RGB image to NOCS. Together with instance mask and depth map, 6D object pose is estimated using shape matching.
Recently, Wang et al.~\cite{wang20196} realized category-level 6D pose tracking based on keypoint matching.

\vspace{6pt}
\paragraph{CASS vs. NOCS}
Although both CASS and NOCS can be regarded as a unified shape space spanning intra-class variations, there are several substantial differences. \emph{First}, NOCS is explicitly defined through consistently aligning all object instances of a category in a normalized 3D space. Our CASS is a shape embedding space implicitly learned with a generative model.
\emph{Second}, when conducting object pose estimation, NOCS is used as the target of pixel-wise correspondence, based on which 6D pose is computed geometrically. In contrast, CASS is treated as a normalized, holistic shape representation from which pose is estimated in an end-to-end and \emph{correspondence-free} manner.
\emph{Third}, different from NOCS where the coordinates are regressed only for visible area, our network learns
to reconstruct a complete 3D shape in CASS which is a global shape understanding beneficial to pose estimation.

% -*- compile-command: "texify --pdf --quiet decop_cvpr18.tex" -*-
% !TEX root = decop_cvpr18.tex
% (shell-command "start decop_cvpr18.pdf")

\section{Model}
Our model is an end-to-end trainable network integrating the learning of both shape space and pose estimation.
%\kx{Special designs are made to match the distribution of the involved two feature spaces, to improve domain transfer.}
We first provide an overview of the network architecture and then elaborate the various network modules.
Training details such as loss functions, parameter setting and training protocol will then follow.

\paragraph{Architecture overview}
\Fig{overview} shows an overview of our network architecture.
The input to the network is a calibrated RGBD image. The one output is the 6DoF pose of the object of interest, represented by a rigid transformation $[R|t]$ with $R\in SO(3)$ and $t\in \mathbb{R}^3$.
\kx{The other output is a reconstructed 3D point cloud of the object in normalized pose but with metric size.}
To handle multiple objects in a cluttered scene, we employ an off-the-shelf object detector to detect and segment the individual object instances.
For each detected object, we crop the RGB image with the bounding box of the segmentation mask and segment it out of the point cloud (converted from the depth image) using the mask, resulting in an image crop and a point patch for the object, respectively.
Both the image crop and the point patch are sent to the main part of our network.

Our core network is composed of three modules responsible for 1) Canonical Shape Space learning, view-factorizing RGBD embedding and point cloud reconstruction, 2) pose-dependent feature extraction and 3) pose estimation, respectively.
The three components are tightly coupled and jointly trained using both synthetic and real-world data.
Next, we elaborate the design of the three modules.
%During test,

\subsection{Canonical Shape Space and View Factorization}
Our goal is to learn a shape space spanning as many shape variants of a category as possible,
where all shapes are pose-normalized but with actual metric size.
Moreover, to work with RGBD inputs, we also need a function to map an RGBG image to
the point in that space representing the corresponding full shape in normalized pose and metric size.
Such a mapping factorizes the view in the RGBD image so that the RGBD feature embedding is view-factorized.
%
%To do that, we learn an auto-encoder of 3D shapes where the bottleneck latent vector serves as the shape embedding. Meanwhile, we need to train a joint embedding of RGBD images and 3D shapes into the shared latent space.

\begin{figure}[t]
\centering
\begin{overpic}
[width=\linewidth]
%[width=\linewidth,grid,tics=10]
{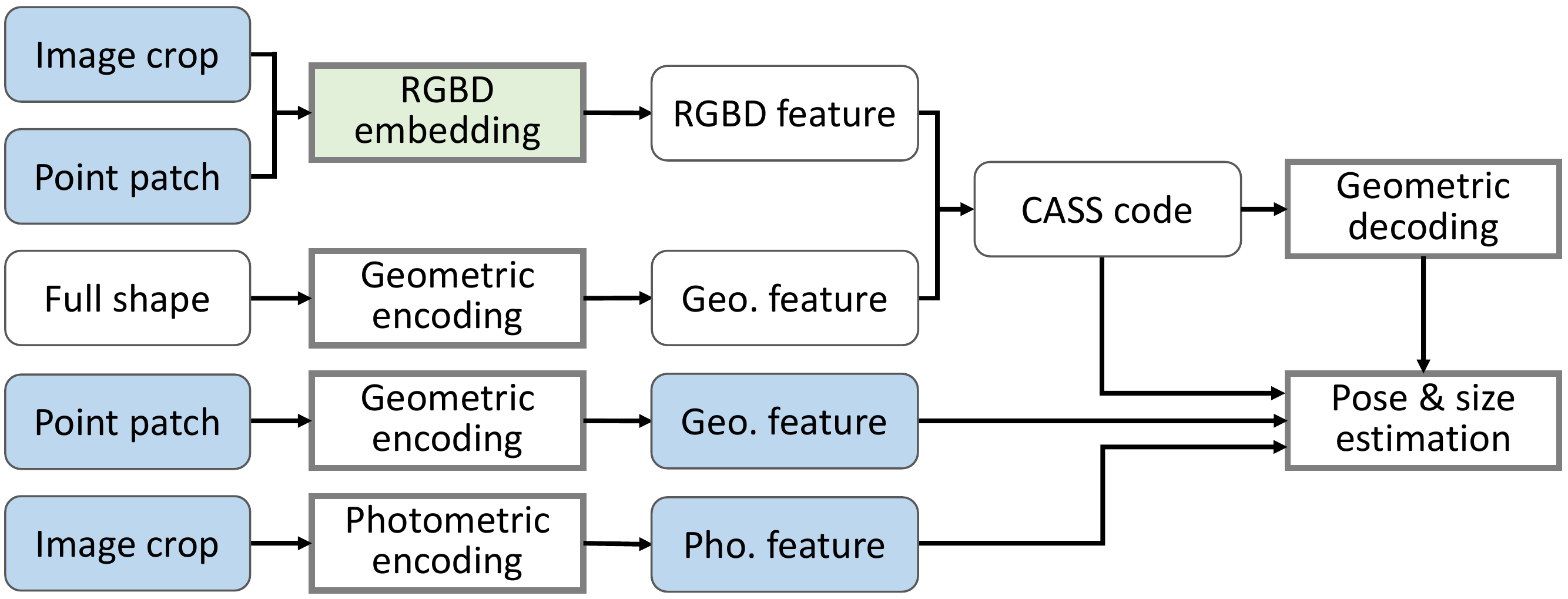}\myfigurename{}
    %\put(17,-3){\small (a)}
    %\put(56,-3){\small (b)}
    %\put(86,-3){\small (c)}
\end{overpic}
\caption{
Illustrating the flow of pose information in our network.
Data and features are depicted with rounded boxes and networks with rectangles.
Data/features shaded in light blue contain pose information; no shading means pose-normalized.
Networks are shaded in light green if they are pose factorizing and no shading otherwise.
}
\label{fig:fac}\vspace{-12pt}
\end{figure}

\paragraph{Learning Canonical Shape Space}
\kx{
We model the space of canonical shapes with the latent space of a deep generative model of pose-normalized shapes. In achieving so, we leverage the publicly available 3D shape repositories such as ShapeNet~\cite{chang2015shapenet}. The 3D models in ShapeNet are consistently oriented and properly scaled within each category.
We sample each model into a point cloud of $M=500$ points. %$2048$
The point sampled 3D shapes, $X_\text{3D}$, are used to train a variational auto-encoder (VAE).
%which adopts PointNet~\cite{qi2017pointnet} as the encoder and FoldingNet~\cite{yang2018foldingnet} as the decoder.
%FoldingNet decoder is capable of recovering the shape point cloud
%through transforming an ellipsoid point cloud using a folding-based transformer network.
The encoder employs the geometric embedding network for 3D point clouds proposed in~\cite{wang2019densefusion}, which is a variant of PointNet~\cite{qi2017pointnet}.
Based on the learned feature, the decoder warps a point cloud of 3D  ellipsoid to match the shape of the input point cloud.
We turn the auto-encoder into a VAE by adding a sampling layer between the encoder and decoder.
The learned posterior distribution $z \sim p(z|X_\text{3D})$ models the space of canonical shapes.
}

\paragraph{Learning view-factorizing RGBD embedding}
Having learned the CASS, our next task is to project an RGBD image in arbitrary view to the space so that the projector functions as view factorization.
%There are many possible solutions of learning cross-modality joint embedding, such as
%feature space alignment with dimension reduction~\cite{li2015joint} and
Such cross-modality data projection task could be finished with the help of data correspondence between the two modalities~\cite{girdhar2016learning}, where metric loss is used to optimize the projector.
Let us refer to this solution as \emph{correspondence-based projection}.
This approach can be adapted to VAE straightforwardly where a projector is trained to map data in one modality to the latent space learned for another modality based on cross-modality data correspondence.
%In the context of VAE, one could learn a joint latent space through training a network to project the data of one modality to the latent space learned for another modality based on cross-modality data correspondence.
However, we found that this method leads to suboptimal point cloud reconstruction and pose estimation due to
1) possibly incorrect correspondences and 2) the compromise between the metric loss and other losses.

To address these issues, we opt for a \emph{joint embedding} approach. Specifically, we learn a VAE which has two encoders mapping both RGBD images and 3D point clouds to a shared latent space. Whilst the 3D encoder adopts PointNet, the RGBD encoder employs the dense fusion architecture proposed in~\cite{wang2019densefusion} (we use the global feature for the whole image instead of the pixel-wise features).
Our crucial design is that the two encoders, albeit having different network architectures,
are trained with \emph{mixed training batch} and \emph{shared training gradients}.
The latter means that gradients computed for either modality are back-propagated to tune both encoders.
Through such mixed training, the learned shared latent space spans the joint feature
space of the both modalities. %Our solution is thus named as \emph{joint-embedding-based projection}.
%See \Fig{fac} for an comparative illustration of the two types of projections.

\begin{figure}[t]
\centering
\begin{overpic}
[width=\linewidth]
%[width=\linewidth,grid,tics=10]
{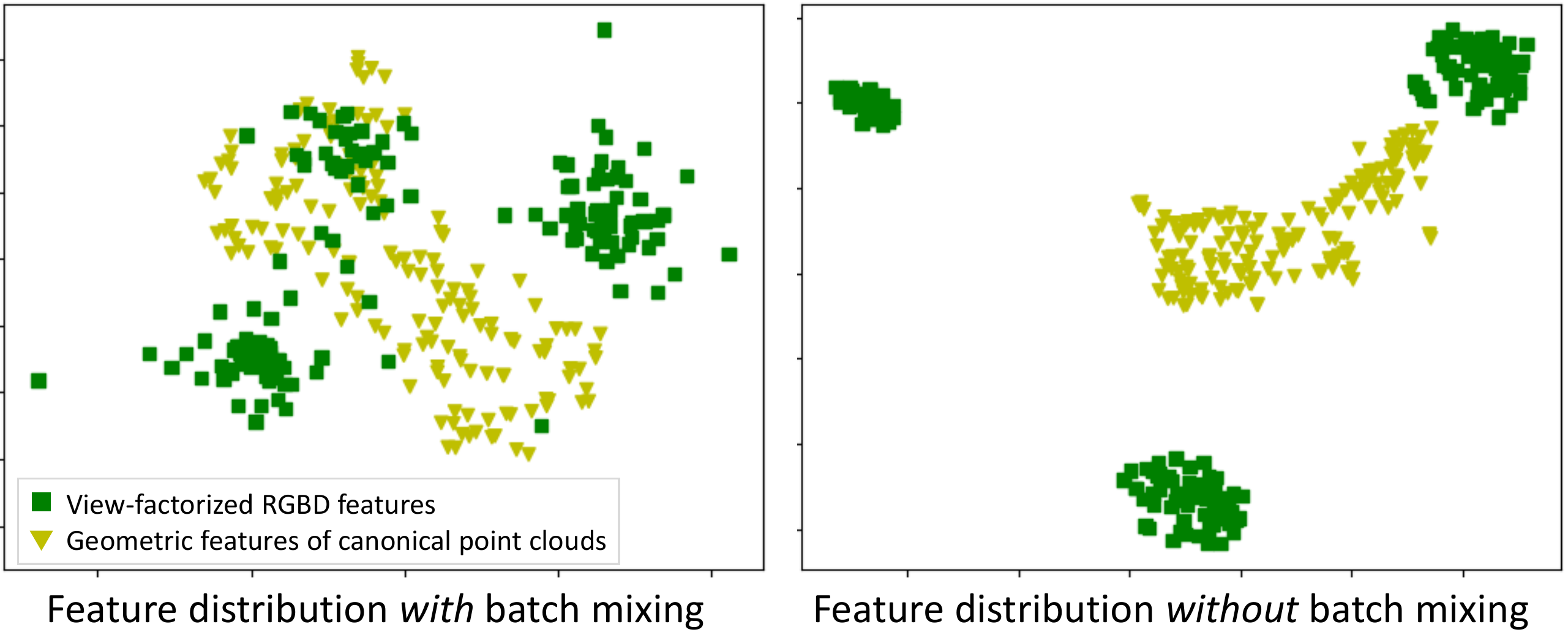}\myfigurename{}
    %\put(17,-3){\small (a)}
    %\put(56,-3){\small (b)}
    %\put(86,-3){\small (c)}
\end{overpic}
\caption{
Comparing the t-SNE plot of view-factorized RGBD features and the geometric features of canonical point clouds for with and without batch mixing. Batch mixing helps to match the distributions of the two features.
}
\label{fig:bm-comp}\vspace{-12pt}
\end{figure}

Compared to correspondence-based approaches, our joint embedding has the following advantages:
\emph{Firstly}, our model can be trained in a correspondence-free or unpaired fashion.
This means the two modalities do not have to share object instances:
It is unnecessary for the object in an RGBD image to have a corresponding 3D model in the training shape set.
\emph{Secondly}, our model introduces no extra loss function other than the basic ones of conventional VAEs.
\emph{Thirdly} and most importantly, the mixed training of the two encoders help to match the feature distributions of the two data modalities (see \Fig{bm-comp}), leading to better model generality and domain transferability.

In summary, the learning of the CASS and the RGBD feature embedding (denoted by $F_\text{vf}$) optimizes the following loss functions:
\begin{equation}\label{eq:loss-cass}
L_\text{CASS} = L_\text{recon}(X_\text{3D},X^\text{R}_\text{3D}) + L_\text{recon}(X_\text{rgbd}^\text{R},X^\ast_\text{3D}) + L_\text{KL},
\end{equation}
where $L_\text{recon}$ and $L_\text{KL}$ are the reconstruction loss and KL divergence loss, respectively.
$X_\text{3D}$ and $X^\text{R}_\text{3D}$ are the input and reconstructed 3D point clouds, and
$X^\text{R}_\text{rgbd}$ and $X^\ast_\text{3D}$ the 3D point cloud reconstructed from the input RGBD and the corresponding ground-truth, respectively.
We use Chamfer distance to measure reconstruction loss.
During test, the 3D point cloud encoder (corresponding to the network branch with red arrows in \Fig{overview}) is discarded and only the RGBD encoder is used for feature extraction.

The RGBD encoder factorizes image view, resulting in pose-independent RGBD feature (CASS code).
\kx{However, the 3D encoder does \emph{not} factorize object pose or size.
This is because both the input and output of the 3D encoder are pose-normalized and metrically sized.
It simply maps a pose-normalized shape to the canonical shape space, without processing on its pose or size.}
\Fig{fac} gives an illustrative summary of the view/pose-factorization ability for all network modules,
and the pose-dependency of all involved data and features.
\Fig{vfc-comp} (top row) shows the t-SNE visualization of the two feature embeddings.
In the plot of view-factorized RGBD features, objects are clustered by category with different poses mixed together, indicating the factorization of view.
The plot of geometric features of canonical point clouds (no pose) also demonstrates category-based clustering effect.

\subsection{Pose-Dependent Feature Extraction}
\begin{figure}[t]
\centering
\begin{overpic}
[width=\linewidth]
%[width=\linewidth,grid,tics=10]
{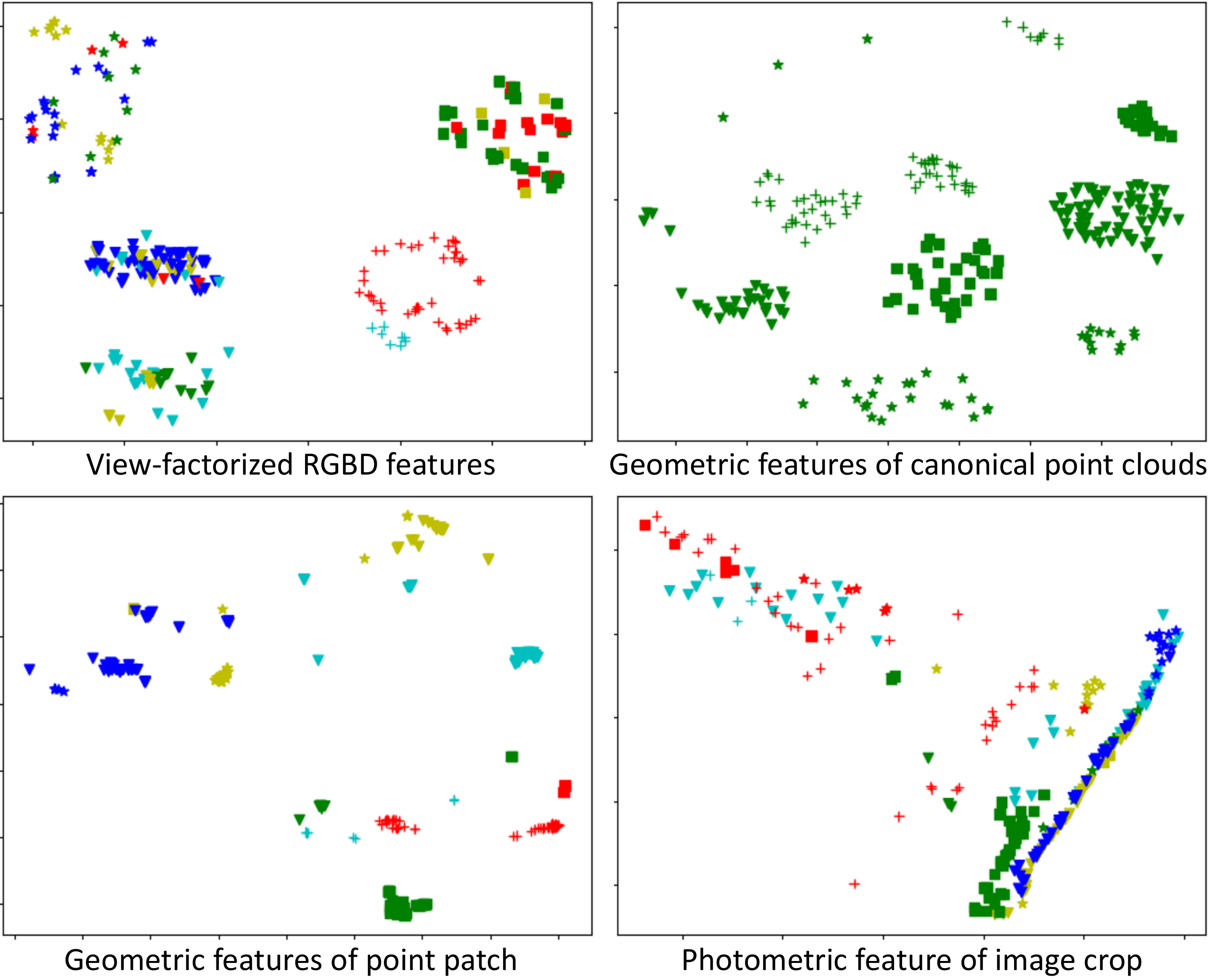}\myfigurename{}
    %\put(17,-3){\small (a)}
    %\put(56,-3){\small (b)}
    %\put(86,-3){\small (c)}
\end{overpic}
\caption{
Comparing t-SNE visualization of the various features involved in our network. Different symbols indicate different object categories while distinct colors correspond to different poses.
}
\label{fig:vfc-comp}\vspace{-12pt}
\end{figure}

To facilitate pose estimation from an input RGBD image, we also extract pose-dependent features for the RGBD image.
We devise two networks for extracting photometric and geometric features separately, based on the RGB and the depth images, respectively.
In our network, these features are used for pose estimation through comparing against pose-dependent features, they are expected to encode the information of pose-color and pose-geometry correlations, respectively.
\Fig{vfc-comp} (bottom row) shows the t-SNE plots of the two features, which both exhibit pose-induced subspace clustering effect.

\paragraph{Photometric feature extraction}
Given the image crop containing the object of interest, we train a fully convolutional network that processes
the color information into a color feature $F_\text{pho}$.
Similar to~\cite{wang2019densefusion},
the image embedding network is an auto-encoder architecture that maps an image of size
$H \times W \times 3$ to a pixel-wise feature map $H \times W \times N$. Each pixel has a $N$-dimensional vector. We then perform an average pooling over all pixel-wise features, obtaining a $N$-dimensional feature for the full image.

\paragraph{Geometric feature extraction}
Given the corresponding point patch, we utilize point-based CNNs to extract an $N$-dim geometric feature $F_\text{geo}$.
\kx{
Here, a key design is that this point-based feature extractor can share the same network of the PointNet-based geometric feature encoder trained for CASS learning. As mentioned above, the geometry encoder is \emph{not} pose-factorizing. Consequently, it can be used to extract pose-dependent geometric features.
Consequently, we have a Siamese network of PointNet-based encoders, one for pose-\emph{independent} CASS embedding and the other for pose-\emph{dependent} geometric feature extraction (see \Fig{overview}).
Having these two tasks share network weights reduces the amount of parameters to be learned. Furthermore, it helps to match the distributions of the CASS codes and the geometric features. This makes them more comparable in facilitating feature-comparison-based pose estimation.}

%\kxc{Draw a figure for illustrating view factorization or not, in according to the overview figure. For example, input has view and output does not, then it is view factorized. If neither input nor output have view, it is not view factorized. If the input has view and the output has too, it is not view factorized too.}

\subsection{Pose and Size Estimation}
We concatenate $F_\text{vf}$, $F_\text{pho}$ and $F_\text{geo}$ into a feature vector of $3N$ length and then feed it into a CNN with 1D convolutions. The output contains a rotation represented by a quaternion $q$ and a 3D translation vector $t$.
\kx{The loss function for pose prediction is defined as the discrepancy between the object point clouds transformed by the ground-truth pose and by the predicted one:
\begin{equation}\label{eq:loss-rts}
%  L_\text{trans} = \|q-q^\ast\|_q + \|t-t^\ast\|_2 + |\frac{s}{s^\ast}-1|,
%  L_\text{trans} = \|q-q^\ast\|_q + \|t-t^\ast\|_2, %+ |\frac{s}{s^\ast}-1|,
   L_\text{pose} = \frac{1}{M}\sum_{i}{\|(Rx_i+t)-(R^\ast x_i+t^\ast)\|},
\end{equation}
%where $q^\ast$, $t^\ast$ and $s^\ast$ are the ground-truth rotation, translation and scaling, respectively. The quaternion norm $\|\cdot\|_q$ is simply the Euclidean norm of the vector of the four quaternion coefficients.
where $x_i$ is the $i$-th point of the $M=500$ sampled points for the object. $[R^\ast|t^\ast]$ and $[R|t]$ are the ground-truth and predicted poses, respectively.}
To handle the alignment ambiguity of symmetric objects, we relax the point-wise matching loss
to Chamfer distance, similar to~\cite{wang2019densefusion}.
Object size is calculated as the dimension of the axis-aligned bounding box (AABB) of the reconstructed 3D point cloud.
%The quaternion norm $\|\cdot\|_q$ is simply the Euclidean norm of the vector of the four quaternion coefficients.}

\subsection{Training Details}

%\paragraph{Training data}
%\kxc{Mainly talk about the preparation of training data. What pre-processing was involved in the preparation?}

\paragraph{Network settings}
The input to our method is a $640\times 480$ RGBD image. With the RGB image, we perform object detection and segmentation. Any off-the-shelf method can be used. For example, we utilize Mask-RCNN~\cite{he2017mask} for the CAMERA dataset.
%, and the method in~\cite{xiang2017posecnn} for the YCB dataset.
%For each input image, we detect object bounding boxes using XXX method. Based on each bounding box, we crop the RGBD image and rescale the crop into a HHHxWWW image.
The image crops do not need to be resize as they are fed into pixel-wise CNNs.
All point patches and 3D models are re-sampled into $500$ points for PointNet feature encoding.
The dimensionality of CASS code and all other features is $N=1024$.
The DenseFusion and FoldingNet modules involved in the various network components use the same network configuration as the original works.
The configuration of all other network modules such as CNNs and MLPs are given in \Fig{overview} (e.g., ``4L'' means four layers and ``1D'' means 1D convolutional layers).
For each convolutional layer in the various modules, we add a Batch Normalization layer followed by an ReLU nonlinearity.
\supl{See more details in the supplemental material.}
%The network used for pose and size estimation contains XXX layers of 1D convolution and YYY layers of MLP.
%group normalization~\cite{wu2018group}

\paragraph{Training protocol}
We adopt a three-stage training.
The \emph{first} stage trains the VAE for CASS learning and view-factorizing RGBD embedding (the part shaded in light blue in \Fig{overview}) for $80$K iterations. The size of mixed batch is $8$ which randomly mixes the training data of RGBD encoding and 3D encoding.
In the \emph{second} stage, we fix the VAE and jointly train pose-dependent feature extraction (the light green part) and pose estimation (the light red part) for $80$K iterations.
The \emph{third} stage then jointly fine-tunes all parts for $40$K iterations.
All training batch has the size of $8$.
%Since the batch size is relatively small, we employ group normalization over batch normalization.
We use an initial learning rate of $0.0001$ and the ADAM optimizer ($\beta_1=0.9$ and $\beta_2=0.999$) with a $1\times 10^{-6}$ weight decay.
In each stage, we decrease the learning rate by a factor of $10$ for every $40$K iterations.

% -*- compile-command: "texify --pdf --quiet decop_cvpr18.tex" -*-
% !TEX root = decop_cvpr18.tex
% (shell-command "start decop_cvpr18.pdf")

%epoch
\section{Results and evaluations}
In this section, we aim to answer the following questions with both qualitative and quantitative evaluations.
1) Whether are the various network modules and design choices necessary?
2) How does our method perform in terms of pose accuracy and when does it outperform the state-of-the-arts?
3) How capable is our network in terms of single-view shape reconstruction?

\subsection{Datasets}
\label{subsec:result-dataset}
We use the datasets from NOCS~\cite{wang2019normalized} which contains six categories: \emph{bottle}, \emph{bowl}, \emph{camera}, \emph{can}, \emph{laptop}, and \emph{mug}. The dataset has two parts: a \emph{real-world dataset} with 4.3K RGBD images from 7 scene videos (3 instances per category) and a \emph{synthetic dataset} with 275K rendered images generated with 1085 model instances from ShapeNetCore~\cite{chang2015shapenet} under random views. We evaluate our method on the NOCS-REAL275 dataset, which contains 2.75K real scene images with 3 unseen instances per category.
In learning the CASS, we also utilized the 3D models from the ShapeNetCore dataset.

\subsection{Evaluation Metrics}
\label{subsec:result-metric}
We follow the evaluation metrics in NOCS~\cite{wang2019normalized} which jointly measure the object detection and pose estimation:
\begin{itemize}
\item \textbf{IoU25 \& IoU50}: the average precision of object instances for which the 3D overlap between the two bounding boxes is larger than $25\%$ or $50\%$ under predicted and ground truth poses respectively;
\item \textbf{5$^\circ$5cm, 10$^\circ$5cm \& 10$^\circ$10cm}: the average precision of object instances for which the the error is less than $n^\circ$ for rotation and $m$ cm for translation. We choose 5$^\circ$5cm, 10$^\circ$5cm, 10$^\circ$10cm similar to~\cite{wang2019normalized}.
\end{itemize}

Additionally, we employ the Chamfer Distance (CD) and Earth Mover's Distance (EMD) to evaluate shape reconstruction from single-view RGBD images.

\subsection{Evaluation on NOCS-REAL275 Dataset}
\label{subsec:result-comp}

%%%%%%%%%%%%%%%%%%%%%
\if 0
\begin{table}[t]
\centering
\caption{Quantitative comparison with NOCS~\cite{wang2019normalized} (we use the 32 bins)}\scalebox{1}{\setlength{\tabcolsep}{1.6mm}{
\begin{tabular}{c|l|l|l|l|l|l}
\hline
\multicolumn{2}{c|}{\multirow{2}{*}{\normalsize{Method}}} & \multicolumn{5}{c}{\small{mAP}}                                                                                                                                              \\ \cline{3-7}
\multicolumn{2}{c|}{}                        & IoU$_{25}$ & IoU$_{50}$ & \begin{tabular}[c]{@{}l@{}}5$^\circ$\\ 5cm\end{tabular} & \begin{tabular}[c]{@{}l@{}}10$^\circ$\\ 5cm\end{tabular} & \begin{tabular}[c]{@{}l@{}}10$^\circ$\\ 10cm\end{tabular} \\ \hline\hline
\multirow{2}{*}{NOCS}           &32 bins             &84.8     &78.0     &10.0     &25.2    &25.8        \\
                                &128 bins            &84.9     &80.5     &9.5      &26.7    &26.7                                                   \\ \hline
\multicolumn{1}{l|}{\multirow{2}{*}{Ours}}   &w/o ICP     &92.3     &86.5     &7.2      &28.9      &29.3                                                   \\
\multicolumn{1}{l|}{}           &with ICP           &100.0   &98.7  &13.0     &37.6     &37.9              \\ \hline
%\multicolumn{2}{c|}{\multirow{1}{*}{Ours}}           &\textbf{100.0}   &\textbf{98.7}  &\textbf{13.0}     &\textbf{37.6}     &\textbf{37.9}              \\ \hline
\end{tabular}}}
\vspace{2pt}
\label{tab:quanti_comp}
\end{table}
\fi
%%%%%%%%%%%%%%%%%%%%%%%%%%%%%%
\begin{table}[t]
\centering
\caption{Quantitative comparison with NOCS~\cite{wang2019normalized} (we use its best-performing variant, i.e., 32-bin NOCS map classification).}\scalebox{1}{\setlength{\tabcolsep}{1.6mm}{
\begin{tabular}{l|c|c|c|c|c}
\hline
\multirow{3}{*}{\normalsize{Method}} & \multicolumn{5}{c}{\small{mAP}}                                                                                                                                              \\ \cline{2-6}
\multicolumn{1}{c|}{}      & IoU$_{25}$ & IoU$_{50}$ & \begin{tabular}[c]{@{}l@{}}5$^\circ$\\ 5cm\end{tabular} & \begin{tabular}[c]{@{}l@{}}10$^\circ$\\ 5cm\end{tabular} & \begin{tabular}[c]{@{}l@{}}10$^\circ$\\ 10cm\end{tabular} \\ \hline\hline
NOCS           &\textbf{84.4}     &\textbf{79.3}     &16.1     &43.7    &43.1        \\ \hline
Ours           &84.2     &77.7     &\textbf{23.5}      &\textbf{58.0}      &\textbf{58.3}      \\ \hline
\end{tabular}}}
\vspace{2pt}
\label{tab:quanti_comp}
\end{table}

\begin{figure}[t]
	\begin{center}
		\includegraphics[width=\linewidth]{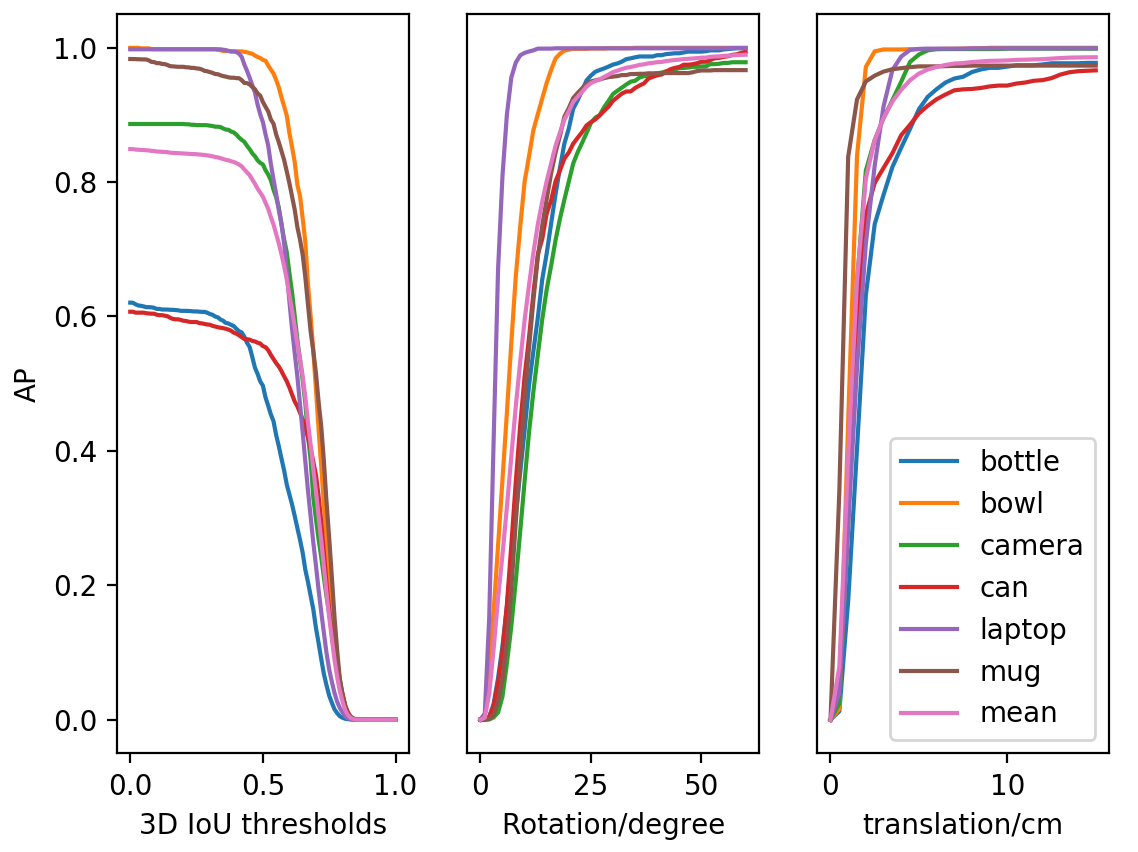}
	\end{center}
    \vspace{-8pt}
	\caption{Result on NOCS REAL275 test dataset, average precision (AP) vs. different thresholds on 3D IoU, rotation error, and translation error.}
    %\vspace{-12pt}
	\label{fig:plot}
\end{figure}

\paragraph{Category-level pose and size estimation}
In Table~\ref{tab:quanti_comp}, we compare our method against NOCS~\cite{wang2019normalized}, which is the state-of-the-art method for category-level 6D object pose and size estimation. In their method, the network is trained to find a normalized coordinate for each pixel and then solve for the pose and size with the help of depth map. On the contrary, our method directly regresses the 6D pose by comparing pose-independent and pose-dependent features. We report the results of NOCS with $32$ pose classification bins, which is its best-performing variant.
\kx{Like NOCS, our results were \emph{not} post-processed, e.g., by ICP refinement, although that is potentially facilitated by our point cloud reconstruction. The results show that our method outperforms NOCS in all metrics except the IoU metrics. The slightly lower IoU values are caused by our less accurate size calculation based on point cloud reconstruction.}
Direct regression of 6D pose is a hard problem. The success of our method is mainly attributed to the strong view-factorized (pose-independent) feature learning with the help of CASS learning and its RGBD embedding. Figure~\ref{fig:plot} shows more detailed analysis with category-wise plots of the various evaluation metrics.

%\begin{itemize}
%  \vspace{-6pt}\item \textbf{B1: Purely Geometric}. Using only depth as input.
%  \vspace{-6pt}\item \textbf{B2: XXX}. XXX.
%  \vspace{-6pt}\item \textbf{C1: NOCS}. XXX.
%\end{itemize}
%Figure~\ref{} or Table~\ref{} shows the comparison ...

\paragraph{Shape reconstruction}
Table~\ref{tab:quanti_recon} reports a quantitative evaluation of 3D point cloud reconstruction from RGBD input, based on the test set of NOCS-REAL275.
From the table, batch mixing leads to much higher reconstruction accuracy in terms of both Chamfer Distance (CD) and Earth Mover's Distance (EMD) metrics. This is because batch mixing ensures the distributions matching between the RGBD embedding and canonical point cloud embedding. This leads to a more accurate RGBD projection (embedding) into the latent shape space.

\begin{table}[t]
\centering
\caption{Evaluation of point cloud reconstruction accuracy with CD ($\times 10^{-3}$) and EMD metrics. The results show that method with batch mixing achieves uniformly higher reconstruction accuracy.}
\scalebox{1}{ %\setlength{\tabcolsep}{2mm}{
\begin{tabular}{l|p{1.2cm}<{\centering}|p{1.2cm}<{\centering}|p{1.2cm}<{\centering}|p{1.2cm}<{\centering}}
\hline
\multirow{2}{*}{} & \multicolumn{2}{c|}{w/o Batch Mixing} & \multicolumn{2}{c}{w/ Batch Mixing}        \\ \cline{2-5}
                  & CD            & EMD           & CD            & EMD                     \\ \hline\hline
bottle            & 1.71           & 0.24          &0.75            &0.04                     \\ \hline
bow               & 0.93           & 0.07          &0.38            &0.04            \\ \hline
camera            & 5.26           & 0.22          &0.77            &0.05            \\ \hline
can               & 1.79           & 0.20          &0.42            &0.04                     \\ \hline
laptop            & 1.94           & 0.10          &3.73            &0.09                     \\ \hline
mug               & 2.40           & 0.11          &0.32            &0.03            \\ \hline
overall           & 2.33           & 0.16          &1.06            &0.05                     \\ \hline
\end{tabular}}
\vspace{2pt}
\label{tab:quanti_recon}
\end{table}

\subsection{Ablation Studies}
\label{subsec:result-ablation}

To experimentally justify the various design choices of our method, we make the following ablations (or their combinations) to our model:
\begin{itemize}
  \vspace{-6pt}\item \textbf{w/o CASS}. Train the pose \& size estimation network without the CASS code as an input.
  \vspace{-6pt}\item \textbf{w/o Distribution Matching (DM)}. Replace the Siamese network with two independent modules, without matching the distribution of the CASS codes and the geometric features.
  \vspace{-6pt}\item \textbf{w/o Batch Mixing (BM)}. Remove batch mixing and use $L_2$ distance as an additional loss to train the projection from an RGBD image in arbitrary view to the canonical shape space.
  %\vspace{-6pt}\item \textbf{No-MC}. No same modality comparison (removing the lowest branch of RGBD encoding).
  \vspace{-6pt}\item \textbf{w/o VAE}. Replacing the VAE with AE.
  %\vspace{-6pt}\item \textbf{BN}. Replacing group normalization layers with batch normalization layers.
\end{itemize}

From the results reported in Table~\ref{tab:ablation_study}, we can see that CASS learning is the most important component for our method. Without CASS learning, the accuracy drops the most especially on the $xx^\circ$$yy$-cm metrics. Next to CASS learning, batch mixing is also very important factors.
\kx{VAE is beneficial to model generalization to unseen objects since it helps learning a well-spanned CASS space with the normal distribution prior. However, it does lead to blurred 3D reconstruction at the same time, which may sacrifices size accuracy (see the IoU comparison in Table~\ref{tab:quanti_comp}). Nevertheless, all factors together contributes the high-precision ($5^\circ$$5$cm) estimation of pose and size.}
%\paragraph{Instance-level pose estimation.}
%Although our method is designed for category-level 6D pose estimation, it can still be used for the instance-level case through .... To gauge the performance of our method in terms of pose accuracy, we compare to the following state-of-the-art methods:
%\begin{itemize}
%  \vspace{-6pt}\item \textbf{I1: DenseFusion}. Using only depth as input.
%  \vspace{-6pt}\item \textbf{I2: XXX}. XXX.
%  \vspace{-6pt}\item \textbf{I3: XXX}. XXX.
%\end{itemize}
%Figure~\ref{} or Table~\ref{} shows the comparison ...

\begin{table}[t]\vspace{10pt}
\centering
\caption{Ablation study of our model. The results show that our full method works the best for most criteria.}
\begin{tabular}{l|l|l|l|l|l|l}
\hline
\multicolumn{2}{c|}{\multirow{3}{*}{\normalsize{Method}}} & \multicolumn{5}{c}{{mAP}}                                                                                                                                              \\ \cline{3-7}
\multicolumn{2}{c|}{}                        & IoU$_{25}$ & IoU$_{50}$ & \begin{tabular}[c]{@{}l@{}}5$^\circ$\\ 5cm\end{tabular} & \begin{tabular}[c]{@{}l@{}}10$^\circ$\\ 5cm\end{tabular} & \begin{tabular}[c]{@{}l@{}}10$^\circ$\\ 10cm\end{tabular} \\ \hline\hline
\multicolumn{1}{l}{}   &w/o CASS          &83.8     &76.2     &4.2      &29.5      &30.0       \\
\multicolumn{1}{l}{}   &w/o BM            &83.6     &77.3     &4.7      &31.8      &32.7       \\
\multicolumn{1}{l}{}   &w/o DM            &84.0     &\textbf{79.0}     &8.4      &39.5      &40.2            \\
\multicolumn{1}{l}{}   &w/o VAE           &83.7     &77.0     &17.0     &42.1      &43.6         \\
\multicolumn{1}{l}{}   &Full           &\textbf{84.2}   &77.7  &\textbf{23.5} &\textbf{58.0}     &\textbf{58.3} \\ \hline
\end{tabular}
\vspace{2pt}
\label{tab:ablation_study}
\end{table}

\subsection{Qualitative results}
\label{subsec:result-qual}

\begin{figure*}[t]
	\begin{center}
		\includegraphics[width=\linewidth]{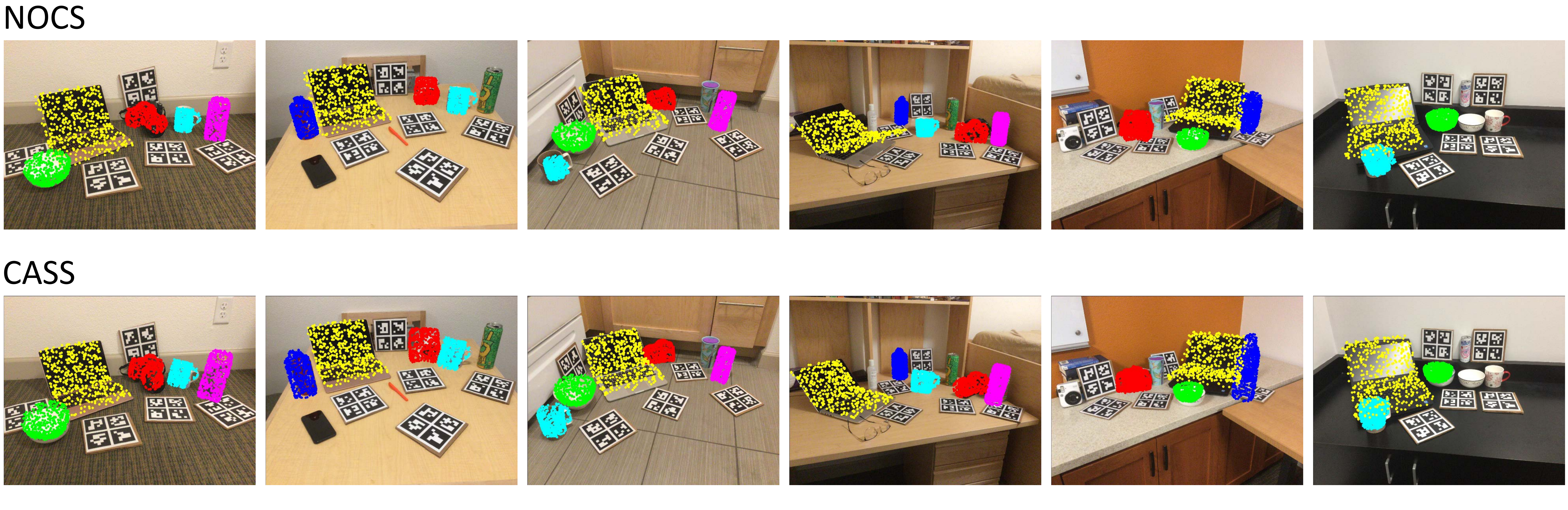}
	\end{center}
    \vspace{-8pt}
	\caption{Qualitative comparison with NOCS~\cite{wang2019normalized} for pose and size estimation (depicted with reconstructed point clouds).}
    %\vspace{-12pt}
	\label{fig:visual_comp}
\end{figure*}

\paragraph{Visual results of pose and size estimation}
Figure~\ref{fig:visual_comp} shows some visual comparisons between our method and NOCS. According to the estimate pose and scale, we draw orientated bounding box for each detected instance overlaid on top of the input RGB images. As can be observed, our method achieves better accuracy especially for size estimation under object occlusion and background distraction.

\begin{figure}[tb]
	\begin{center}
		\includegraphics[width=\linewidth]{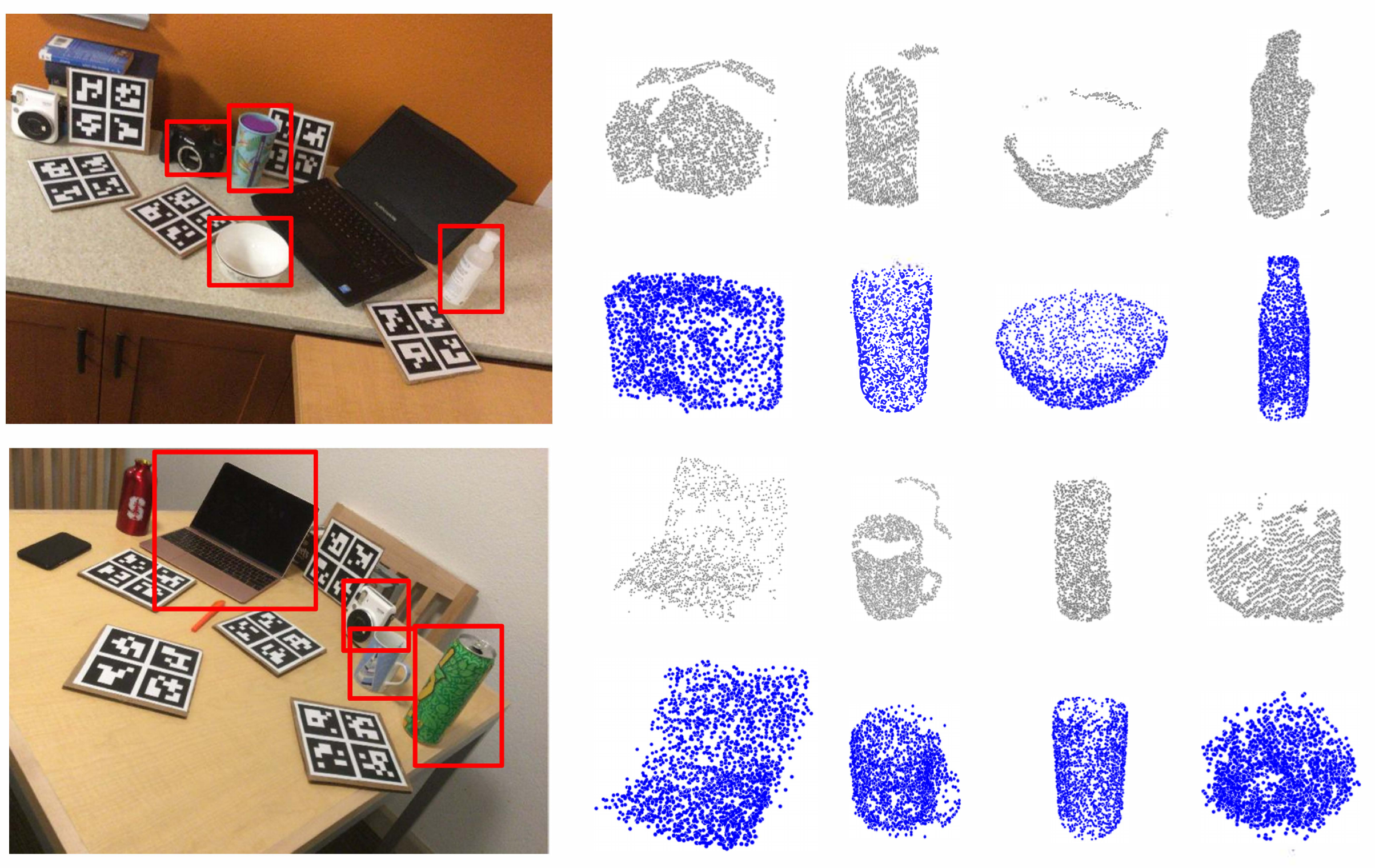}
	\end{center}
    \vspace{-8pt}
	\caption{3D Reconstruction from single-view RGBD. The point clouds in grey color are unprojected from depth maps. The blue point clouds are reconstructed by our network.}
    %\vspace{-12pt}
	\label{fig:recon}
\end{figure}

\paragraph{Visual results of shape reconstruction}
Figure~\ref{fig:recon} shows visual results of shape reconstruction. Our method is able to reconstruct full 3D shapes in point cloud from single-view RGBD images, contrasting them with the point clouds unprojected from depth maps.

%Show a few examples that can recover full shapes from partially observed shapes so that pose estimation can be accurate for cases with severe occlusion

%\paragraph{Visualization of VFASS.}
%t-SNE of the learned shape embedding space

\section{Conclusion}

We have presented a novel correspondence-free approach to category-level object pose and size estimation.
This is achieved by learning a shape space of 3D models in normalized pose and metric size
based on deep generative model.
The input RGBD image is embedded into the shape space, extracting pose-independent features.
Pose estimation is realized by comparing the pose-independent and pose-dependent features.
Evaluation shows that our method arrives at the state-of-the-art performance.

\vspace{6pt}
\paragraph{Limitations and future work}
Our current method has a few limitations on which we aim to improve as future work.
\emph{First}, our method cannot handle well very complex shapes due to the difficulty in reconstructing shapes with complicated geometry (e.g. high genus). In this aspect, our method can be enhanced by learning a more powerful shape reconstruction with, e.g., volumetric 3D representation.
\emph{Second}, our current method does not close the loop in terms of utilizing the reconstructed shape geometry to guide/supervise the training of pose estimation. This may lead to a unsupervised or self-taught approach which we plan to investigate in a future work.
\emph{Third}, our method still cannot achieve very high precision, as reflected by the relatively lower accuracy for the $5^\circ$$5$cm metric. This may be an inherent limitation for a correspondence-free or sparse approach.
Note, however, our method did not use ICP to refine the pose or size.
\emph{Last}, we plan to extend our current framework to online object pose tracking similar to~\cite{wang20196}.

\section*{Acknowledgement}
We thank the anonymous reviewers for the valuable suggestions.
We are grateful to Chen Wang, one of the authors of DenseFusion, for the help and discussion.
This work was supported in part by the National Key Research and Development Program of China (No. 2018AAA0102200), the NSFC (61572507, 61532003, 61622212, 61902419), the NUDT Research Grants (No.ZK19-30) and the Natural Science Foundation of Hunan Province for Distinguished Young Scientists (2017JJ1002).

{\small
\bibliographystyle{ieee_fullname}
\bibliography{6dpose}

\begin{thebibliography}{10}\itemsep=-1pt

\bibitem{avetisyan2019scan2cad}
Armen Avetisyan, Manuel Dahnert, Angela Dai, Manolis Savva, Angel~X Chang, and
  Matthias Nie{\ss}ner.
\newblock Scan2cad: Learning cad model alignment in rgb-d scans.
\newblock In {\em Proceedings of the IEEE Conference on Computer Vision and
  Pattern Recognition}, pages 2614--2623, 2019.

\bibitem{balntas2017pose}
Vassileios Balntas, Andreas Doumanoglou, Caner Sahin, Juil Sock, Rigas
  Kouskouridas, and Tae-Kyun Kim.
\newblock Pose guided rgbd feature learning for 3d object pose estimation.
\newblock In {\em Proceedings of the IEEE International Conference on Computer
  Vision}, pages 3856--3864, 2017.

\bibitem{brachmann2014learning}
Eric Brachmann, Alexander Krull, Frank Michel, Stefan Gumhold, Jamie Shotton,
  and Carsten Rother.
\newblock Learning 6d object pose estimation using 3d object coordinates.
\newblock In {\em European conference on computer vision}, pages 536--551.
  Springer, 2014.

\bibitem{chang2015shapenet}
Angel~X Chang, Thomas Funkhouser, Leonidas Guibas, Pat Hanrahan, Qixing Huang,
  Zimo Li, Silvio Savarese, Manolis Savva, Shuran Song, Hao Su, et~al.
\newblock Shapenet: An information-rich 3d model repository.
\newblock {\em arXiv preprint arXiv:1512.03012}, 2015.

\bibitem{choi20123d}
Changhyun Choi and Henrik~I Christensen.
\newblock 3d pose estimation of daily objects using an rgb-d camera.
\newblock In {\em 2012 IEEE/RSJ International Conference on Intelligent Robots
  and Systems}, pages 3342--3349. IEEE, 2012.

\bibitem{choi2012voting}
Changhyun Choi, Yuichi Taguchi, Oncel Tuzel, Ming-Yu Liu, and Srikumar
  Ramalingam.
\newblock Voting-based pose estimation for robotic assembly using a 3d sensor.
\newblock In {\em 2012 IEEE International Conference on Robotics and
  Automation}, pages 1724--1731. IEEE, 2012.

\bibitem{do2018deep}
Thanh-Toan Do, Ming Cai, Trung Pham, and Ian Reid.
\newblock Deep-6dpose: Recovering 6d object pose from a single rgb image.
\newblock {\em arXiv preprint arXiv:1802.10367}, 2018.

\bibitem{georgakis2018matching}
Georgios Georgakis, Srikrishna Karanam, Ziyan Wu, and Jana Kosecka.
\newblock Matching rgb images to cad models for object pose estimation.
\newblock {\em arXiv preprint arXiv:1811.07249}, 2018.

\bibitem{girdhar2016learning}
Rohit Girdhar, David~F Fouhey, Mikel Rodriguez, and Abhinav Gupta.
\newblock Learning a predictable and generative vector representation for
  objects.
\newblock In {\em European Conference on Computer Vision}, pages 484--499.
  Springer, 2016.

\bibitem{he2017mask}
Kaiming He, Georgia Gkioxari, Piotr Doll{\'a}r, and Ross Girshick.
\newblock Mask r-cnn.
\newblock In {\em Proceedings of the IEEE international conference on computer
  vision}, pages 2961--2969, 2017.

\bibitem{hodavn2015detection}
Tom{\'a}{\v{s}} Hoda{\v{n}}, Xenophon Zabulis, Manolis Lourakis,
  {\v{S}}t{\v{e}}p{\'a}n Obdr{\v{z}}{\'a}lek, and Ji{\v{r}}{\'\i} Matas.
\newblock Detection and fine 3d pose estimation of texture-less objects in
  rgb-d images.
\newblock In {\em Proc. IROS}, pages 4421--4428. IEEE, 2015.

\bibitem{izadinia2017im2cad}
Hamid Izadinia, Qi Shan, and Steven~M Seitz.
\newblock Im2cad.
\newblock In {\em Proceedings of the IEEE Conference on Computer Vision and
  Pattern Recognition}, pages 5134--5143, 2017.

\bibitem{kehl2017ssd}
Wadim Kehl, Fabian Manhardt, Federico Tombari, Slobodan Ilic, and Nassir Navab.
\newblock {SSD-6D}: Making rgb-based 3d detection and 6d pose estimation great
  again.
\newblock In {\em Proceedings of the IEEE International Conference on Computer
  Vision}, pages 1521--1529, 2017.

\bibitem{kehl2016deep}
Wadim Kehl, Fausto Milletari, Federico Tombari, Slobodan Ilic, and Nassir
  Navab.
\newblock Deep learning of local rgb-d patches for 3d object detection and 6d
  pose estimation.
\newblock In {\em European Conference on Computer Vision}, pages 205--220.
  Springer, 2016.

\bibitem{konishi2018real}
Yoshinori Konishi, Kosuke Hattori, and Manabu Hashimoto.
\newblock Real-time 6d object pose estimation on cpu.
\newblock {\em arXiv preprint arXiv:1811.08588}, 2018.

\bibitem{li2018deepim}
Yi Li, Gu Wang, Xiangyang Ji, Yu Xiang, and Dieter Fox.
\newblock Deepim: Deep iterative matching for 6d pose estimation.
\newblock In {\em Proceedings of the European Conference on Computer Vision
  (ECCV)}, pages 683--698, 2018.

\bibitem{peng2019pvnet}
Sida Peng, Yuan Liu, Qixing Huang, Xiaowei Zhou, and Hujun Bao.
\newblock {PVNet}: Pixel-wise voting network for 6dof pose estimation.
\newblock In {\em Proc. CVPR}, pages 4561--4570, 2019.

\bibitem{qi2017pointnet}
Charles~R Qi, Hao Su, Kaichun Mo, and Leonidas~J Guibas.
\newblock Pointnet: Deep learning on point sets for 3d classification and
  segmentation.
\newblock In {\em Proceedings of the IEEE Conference on Computer Vision and
  Pattern Recognition}, pages 652--660, 2017.

\bibitem{rad2017bb8}
Mahdi Rad and Vincent Lepetit.
\newblock Bb8: A scalable, accurate, robust to partial occlusion method for
  predicting the 3d poses of challenging objects without using depth.
\newblock In {\em Proc. ICCV}, pages 3828--3836, 2017.

\bibitem{sahin2019instance}
Caner Sahin, Guillermo Garcia-Hernando, Juil Sock, and Tae-Kyun Kim.
\newblock Instance-and category-level 6d object pose estimation.
\newblock {\em arXiv preprint arXiv:1903.04229}, 2019.

\bibitem{sahin2018category}
Caner Sahin and Tae-Kyun Kim.
\newblock Category-level 6d object pose recovery in depth images.
\newblock In {\em Proceedings of the ECCV Workshop}, 2018.

\bibitem{sahin2016iterative}
Caner Sahin, Rigas Kouskouridas, and Tae-Kyun Kim.
\newblock Iterative hough forest with histogram of control points for 6 dof
  object registration from depth images.
\newblock In {\em Proc. IROS}, pages 4113--4118. IEEE, 2016.

\bibitem{sock2017multi}
Juil Sock, S Hamidreza~Kasaei, Luis Seabra~Lopes, and Tae-Kyun Kim.
\newblock Multi-view 6d object pose estimation and camera motion planning using
  rgbd images.
\newblock In {\em Proceedings of the IEEE International Conference on Computer
  Vision}, pages 2228--2235, 2017.

\bibitem{sock2018multi}
Juil Sock, Kwang~In Kim, Caner Sahin, and Tae-Kyun Kim.
\newblock Multi-task deep networks for depth-based 6d object pose and joint
  registration in crowd scenarios.
\newblock {\em arXiv preprint arXiv:1806.03891}, 2018.

\bibitem{tejani2014latent}
Alykhan Tejani, Danhang Tang, Rigas Kouskouridas, and Tae-Kyun Kim.
\newblock Latent-class hough forests for 3d object detection and pose
  estimation.
\newblock In {\em Proc. ECCV}, pages 462--477. Springer, 2014.

\bibitem{tekin2018real}
Bugra Tekin, Sudipta~N Sinha, and Pascal Fua.
\newblock Real-time seamless single shot 6d object pose prediction.
\newblock In {\em Proceedings of the IEEE Conference on Computer Vision and
  Pattern Recognition}, pages 292--301, 2018.

\bibitem{wang20196}
Chen Wang, Roberto Mart{\'\i}n-Mart{\'\i}n, Danfei Xu, Jun Lv, Cewu Lu, Li
  Fei-Fei, Silvio Savarese, and Yuke Zhu.
\newblock 6-pack: Category-level 6d pose tracker with anchor-based keypoints.
\newblock {\em arXiv preprint arXiv:1910.10750}, 2019.

\bibitem{wang2019densefusion}
Chen Wang, Danfei Xu, Yuke Zhu, Roberto Mart{\'\i}n-Mart{\'\i}n, Cewu Lu, Li
  Fei-Fei, and Silvio Savarese.
\newblock Densefusion: 6d object pose estimation by iterative dense fusion.
\newblock In {\em Proc. CVPR}, pages 3343--3352, 2019.

\bibitem{wang2019normalized}
He Wang, Srinath Sridhar, Jingwei Huang, Julien Valentin, Shuran Song, and
  Leonidas~J Guibas.
\newblock Normalized object coordinate space for category-level 6d object pose
  and size estimation.
\newblock In {\em Proc. CVPR}, pages 2642--2651, 2019.

\bibitem{wohlhart2015learning}
Paul Wohlhart and Vincent Lepetit.
\newblock Learning descriptors for object recognition and 3d pose estimation.
\newblock In {\em Proceedings of the IEEE Conference on Computer Vision and
  Pattern Recognition}, pages 3109--3118, 2015.

\bibitem{xiang2017posecnn}
Yu Xiang, Tanner Schmidt, Venkatraman Narayanan, and Dieter Fox.
\newblock Posecnn: A convolutional neural network for 6d object pose estimation
  in cluttered scenes.
\newblock {\em arXiv preprint arXiv:1711.00199}, 2017.

\bibitem{xiao2019pose}
Yang Xiao, Xuchong Qiu, Pierre-Alain Langlois, Mathieu Aubry, and Renaud
  Marlet.
\newblock Pose from shape: Deep pose estimation for arbitrary 3d objects.
\newblock {\em arXiv preprint arXiv:1906.05105}, 2019.

\bibitem{zhang2019holistic}
Haoruo Zhang and Qixin Cao.
\newblock Holistic and local patch framework for 6d object pose estimation in
  rgb-d images.
\newblock {\em Computer Vision and Image Understanding}, 180:59--73, 2019.

\end{thebibliography}
}

\end{document}